%% file: main.tex
\pdfoutput=1

\documentclass[11pt]{article}

\usepackage[]{acl}

\usepackage{times}
\usepackage{latexsym}

\usepackage[T1]{fontenc}

\usepackage[utf8]{inputenc}

\usepackage{microtype}

\usepackage{inconsolata}

\usepackage{graphicx}
\usepackage{url}
\usepackage{latexsym}
\usepackage{makecell}
\usepackage{epstopdf}
\usepackage{tipa}
\usepackage{hyperref}
\usepackage{xstring}
\usepackage{amsfonts}
\usepackage{amsmath}
\usepackage{fixltx2e}
\usepackage[T1]{fontenc}
\usepackage{booktabs}
\usepackage{arabtex}
\usepackage{hhline}
\usepackage{pdfpages}
\usepackage{utf8}
\input{macros.tex}
\usepackage{multirow}
\usepackage{subfigure}
\usepackage{amssymb}
\usepackage[super]{nth}
\usepackage{arydshln}

\usepackage{lipsum}
\usepackage{bm}

\usepackage{caption}
\setcode{utf8}
\vocalize
\usepackage{xcolor}

\newcommand{\camelmorph}{{\sc CamelMorph}}

\title{Computational Morphology and Lexicography Modeling \\ of Modern Standard Arabic Nominals}

\author{
Christian Khairallah,\textsuperscript{\textdagger}
Reham Marzouk,\textsuperscript{\textdagger,\textdagger\textdagger}
Salam Khalifa,\textsuperscript{\textdagger,\textdaggerdbl}\\
{\bf Mayar Nassar,\textsuperscript{\textdagger,\textdaggerdbl\textdaggerdbl}
Nizar Habash\textsuperscript{\textdagger}}\\
Computational Approaches to Modeling Language (CAMeL) Lab\\  \textsuperscript{\textdagger}New York University Abu Dhabi,
\textsuperscript{\textdagger\textdagger}Alexandria University\\
\textsuperscript{\textdaggerdbl}Stony Brook University, \textsuperscript{\textdaggerdbl\textdaggerdbl}Ain Shams University\\
{\small \texttt{\{christian.khairallah,nizar.habash\}@nyu.edu},
\texttt{igsr.r.marzouk@alexu.edu.eg},}\\
{\small \texttt{salam.khalifa@stonybrook.edu}, \texttt{mayar.nassar@art.asu.edu.eg}
}}

\begin{document}
\maketitle
\begin{abstract}
Modern Standard Arabic (MSA) nominals present many morphological and lexical modeling challenges that have not been consistently addressed previously. 
This paper attempts to define the space of such challenges, and 
leverage a recently proposed morphological framework to build a comprehensive and extensible model for MSA nominals.  Our model design addresses the nominals' intricate morphotactics, as well as their paradigmatic irregularities. Our implementation showcases enhanced accuracy and consistency compared to a commonly used MSA morphological analyzer and generator. We make our models publicly available.
\end{list}
\end{abstract}

\section{Introduction}
\label{sec:intro}


Arabic poses many challenges to computational morphology: its hybrid
templatic and concatenative processes, rich collections of inflectional and cliticization features, numerous allomorphs, and highly ambiguous orthography.
Over the decades, many approaches have been explored in developing Arabic morphological analyzers and generators \cite{Beesley:1989:two-level,Kiraz:1994:multi-tape,Buckwalter:2004:buckwalter,Graff:2009:standard,habash-etal-2022-morphotactic}.
These tools continue to show value for Arabic natural language processing (NLP) even when paired with state-of-the-art neural models on various tasks such as  morphological tagging \cite{Zalmout:2017:dont,inoue-etal-2022-morphosyntactic}, sentiment analysis \cite{Baly:2017:sentiment}, and controlled text rewriting \cite{alhafni-etal-2022-user}.
Developing such tools is neither cheap nor easy; and some of them are not freely available, or incomplete, e.g., \newcite{habash-etal-2022-morphotactic} points out how a popular Arabic analyzer, SAMA \cite{Graff:2009:standard}, has very low coverage for phenomena such as command form or passive voice.  

The effort presented in this paper is about the modeling of Modern Standard Arabic (MSA) nominals in an open-source Arabic morphology project ({\camelmorph}) introduced by \newcite{habash-etal-2022-morphotactic}, who demonstrated their approach on verbs in MSA and Egyptian Arabic.  Verbs are generally seen as the \textit{sweethearts} of Arabic computational morphology: while they have some complexity, they are very regular and predictable.  Nominals are far more complex --- in addition to their numerous morphotactics, they have complicated paradigms with different degrees of completeness and many irregular forms, e.g., broken plural and irregular feminines \cite{Alkuhlani:2011:corpus}.

Our contributions are (a) \textbf{defining} the space of challenges in modeling MSA nominals (\textit{nouns}, \textit{adjectives}, and \textit{elatives/comparative adjectives}); 
(b)~\textbf{developing} a large-scale implementation which is easily extendable within the recently introduced {\camelmorph} framework;  (c) \textbf{benchmarking} our models against a popular Arabic morphology database \cite{Graff:2009:standard,Taji:2018:arabic-morphological} and demonstrating them to be more accurate and consistent; and finally (d) making our databases and code \textbf{publicly available}.\footnote{All system details and guidelines are available under the \texttt{official\_releases/eacl2024\_release/} directory of the project GitHub: \url{http://morph.camel-lab.com}.\label{github}}

Next, we present relevant terminology (\S\ref{sec:terminology}), and related work (\S\ref{sec:related-work}). We follow with a discussion of Arabic nominal modeling challenges (\S\ref{sec:challenges}), and give an overview on the {\camelmorph} framework (\S\ref{sec:camel-morph-approach}) and how we utilize it (\S\ref{sec:nominals-camel-morph}). Finally, we present an evaluation of our system (\S\ref{sec:evaluation}).

\begin{table*}[t!]
\centering
 \includegraphics[width=1\textwidth]{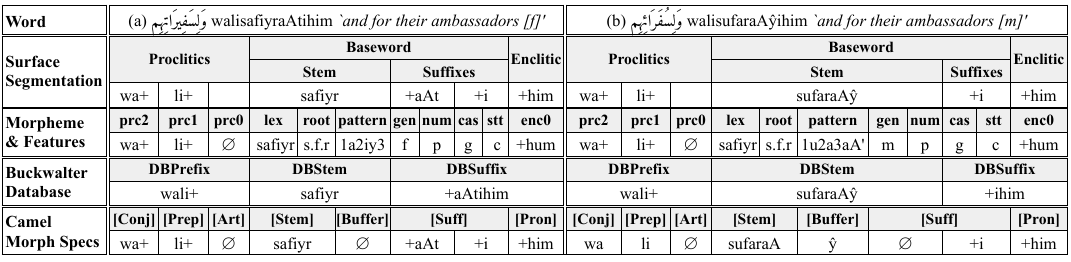}
    \caption{Two examples in four different Arabic morphological representation schemes.}
\label{tokenization-table}
\end{table*}

\section{Relevant Terminology}
\label{sec:terminology}

We present the relevant terminology we use in this paper and illustrate it with examples in Table~\ref{tokenization-table}. The table presents four different ways to represent the morphological information. 
Arabic words are created by combining different types of \textbf{morphemes}: some are concatenative \textbf{affixes} (\textit{nominals only take \textbf{suffixes}}) and \textbf{clitics}, and others are templatic \textbf{roots} and \textbf{patterns} that interdigitate to form \textbf{stems}, which concatenate with the suffixes and clitics. Nominal suffixes typically represent \textbf{gender}, \textbf{number}, \textbf{case} and \textbf{state} features. However, occasionally some of these features are realized through patterns, e.g., Table~\ref{tokenization-table}~(b)'s example of \textbf{templatic} (aka \textbf{broken}) \textbf{plural}.
\textbf{Proclitics} (conjunctions, prepositions, and definite article) and \textbf{enclitics} (possessive pronouns) are syntactically independent but phono-orthographically dependent morphemes.
We use the term \textbf{baseword} to refer to the most basic complete word form (stem+suffixes) without clitics. 
Some morphemes have contextually variable alternatives, called \textbf{allomorphs}, e.g., in Table~\ref{tokenization-table}, the enclitic \<هُم>+ {\textit{+hum}}\footnote{HSB Arabic transliteration \cite{Habash:2007:arabic-transliteration}.} has an allomorph  \<هِم>+ {\textit{+him}} which is used if an /i/ vowel precedes it.  
Systematic allomorphic changes in stem endings can be represented using stem sub-strings called \textbf{stem buffers} \cite{habash-etal-2022-morphotactic}, e.g., Table~\ref{tokenization-table}~(b)'s \textbf{[Buffer]} in the Camel Morph Specs row has two other forms that may vary based on the vowel of the suffix that follows it:
(\<ئ>\texttt{|}\<ؤ>\texttt{|}\<ء>)\<سُفَرَا> \textit{sufaraA('|{\WHAMZA}|{\YHAMZA})}.

At a higher level beyond a single word, and inspired by  \newcite{Stump:2001:inflectional},  we define the  \textbf{lexeme} as the set of words varying through inflection and cliticization operations. The lexeme is headed by a representative form called the \textbf{lemma} (\textbf{lex} in Table~\ref{tokenization-table}). We refer to the \textbf{paradigm} as the space occupied by a lexeme over the inflectional grid, which is structured according to a set of morphosyntactic \textbf{functional features}. Different combinations of the \textbf{values} of these features define \textbf{paradigm slots}, and these slots are either occupied by one word form or more (e.g., words having two plural forms), or they may be empty. For an Arabic nominal, the obligatory features are \textbf{POS}, \textbf{case}, \textbf{state}, \textbf{gender}, and \textbf{number}, and optional ones come in the form of concatenative \textbf{clitics} \cite{Habash:2010:introduction}. Hence, given a lemma and a set of feature values, one can generate all the word forms in a lexeme, i.e., \textbf{inflection}. Within this framework, any other (i.e., non-inflectional) morphological transformation maintaining the same templatic root of a lexeme results in a different lexeme, and this is called \textbf{derivation}.

Finally, Appendix \ref{glossary} presents a glossary of the discussed terms, with their abbreviations,\footnote{A quick reference to abbreviations: \underline{\textbf{m}}asculine, \underline{\textbf{f}}eminine, \underline{\textbf{s}}ingular, and \underline{\textbf{p}}lural for functional gender-number; \underline{\textbf{M}}asculine, \underline{\textbf{F}}eminine, \underline{\textbf{S}}ingular, and \underline{\textbf{P}}lural for form gender-number; \underline{\textbf{a}}ccusative, \underline{\textbf{n}}ominative, and \underline{\textbf{g}}enitive for case; \underline{\textbf{i}}ndefinite, \underline{\textbf{d}}efinite, and \underline{\textbf{c}}onstruct for state; \underline{\textbf{1}} and \underline{\textbf{3}} for 1\textsuperscript{st} and 3\textsuperscript{rd} person; \underline{\textbf{N}}oun (\underline{\textbf{R}}ational or \underline{\textbf{I}}rrational) or \underline{\textbf{A}}djective for POS. \label{abbreviations}} Arabic equivalents, and examples.

\section{Related Work}
\label{sec:related-work}


\paragraph{Morphological Analysis \& Generation}
This work builds on a long history of morphological analysis and generation tools which may, or may not, have tried to extensively model Arabic nominals \cite{Al-Sughaiyer:2004:arabic,habash:2007:representations,sawalha:2008:comparative,Habash:2010:introduction,Altantawy:2011:fast}. \newcite{Altantawy:2011:fast} categorizes different approaches along a conituum based on their  modeling of morphological representations of words. At one end, the representations are characterized by rich linguistic abstractions and a greater reliance on a templatic-affixational perspective of morphology \cite{Beesley:1989:two-level,Kiraz:1994:multi-tape,Beesley:1996:arabic,Habash:2006:magead,Smrvz:2007:elixirfm,Boudchiche:2017:alkhalil};
while at the other end, the representations tend to be more surface-form oriented and organized along precompiled derivation-inflectional solutions
\cite{Buckwalter:2004:buckwalter,Graff:2009:standard,Taji:2018:arabic-morphological}. The former tends to rely on multi-tiered representations that map underlying forms to surface forms, generally using  finite-state transducers through complex rules; and can either model at the morpheme \cite{Beesley:1996:arabic} or lexeme level \cite{Smrvz:2007:elixirfm}. The latter tends to follow a more stem-based approach where morphotactic rules are built directly into the lexicon and inherently models at the morpheme and features level, without including roots and patterns into the rules. The most widely used of these models rely on the six-table approach used in the Buckwalter/Standard Arabic Morphological Analyzer (BAMA/SAMA) \cite{Buckwalter:2004:buckwalter,Graff:2009:standard}, which entails a lexicon of morphemes and compatibility tables between them.

Aligned, to a degree, with the stem-based methodologies, \newcite{habash-etal-2022-morphotactic} presented a \textit{middle ground} approach, within the open-source Arabic morphology project {\camelmorph}. They modeled morphotactic allomorphy  via linguistically motivated inter-allomorphic compatibility rules, and facilitated the creation of lexicons (closed and open-class) that are comparatively easy to manipulate and modify.
They demonstrated their approach building on top of, and comparing to, \newcite{Buckwalter:2004:buckwalter}'s latest extension \cite{Taji:2018:arabic-morphological}. They presented results on modeling the Arabic verbal system in MSA and Egyptian Arabic.  In this paper, we leverage their approach to comprehensively model MSA nominals.

\paragraph{Computational Modeling of Arabic Nominals}

Modeling Arabic nominal morphology presents a more intricate challenge when compared to verbs, as the latter generally follow strictly regular inflectional patterns \cite{Al-Sughaiyer:2004:arabic,Altantawy:2010:morphological,Habash:2010:introduction,Alkuhlani:2011:corpus}. Even when nominals are modeled, their treatment is often incomplete. For example, broken plurals are not always linked to their singular forms (or lemmas), which adds a cost to using them in downstream applications \cite{Xu:2002:empirical}.
Even in systems that modeled  broken plurals lexically, e.g., \newcite{Buckwalter:2004:buckwalter}, 
there were major gaps such as not specifying their functional gender and number \cite{Smrvz:2007:functional,Alkuhlani:2011:corpus}.  
Furthermore, \newcite{Buckwalter:2004:buckwalter} confounded the definite and construct states for some morphemes \cite{Smrvz:2007:functional}. 

Several attempts were undertaken to tackle these issues \cite{Soudi:2001:computational,Smrvz:2007:functional,Habash:2007:determining,Altantawy:2010:morphological,Alkuhlani:2011:corpus,neme:2013:pattern,Taji:2018:arabic-morphological}; however, they either lacked a comprehensive approach, focused only on a subset of nominals, or proved challenging to extend straightforwardly.

\section{Arabic Nominal Morphology}
\label{sec:challenges}
Default word composition assumes a straightforward one-to-one mapping from features to morphemes, with simple interdigitation and concatenation.  In practice, however, there are many variations and exceptions. We outline the most important issues next, starting with word-level inflection and cliticization, and following with lexicographic and paradigmatic challenges. 

\begin{table}[t!]
\centering
 \includegraphics[width=0.48\textwidth]{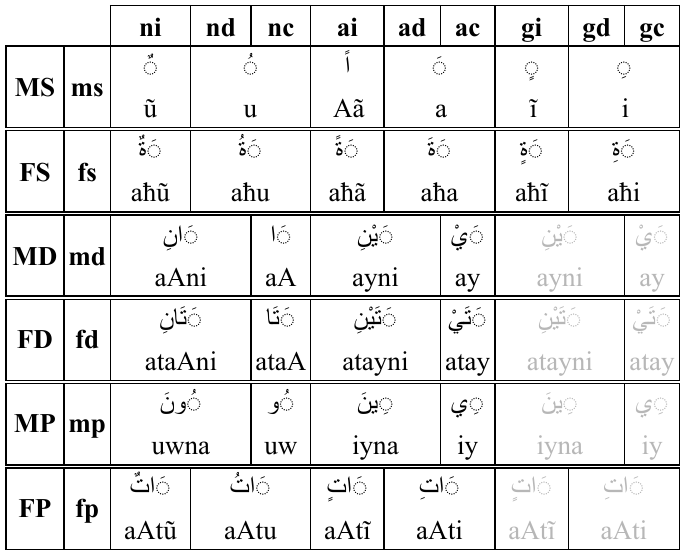}
    \caption{The set of MSA nominal suffixes and their \textbf{default} mapping to functional values of gender-number (rows) and case-state (columns). The capitalized tags refer to the set of suffixes by form, not function. Trivially, they match here because this is a \textit{default mapping table}.
    Merged cells indicate instances of syncretism in adjacent  cells. Greyed cells indicate syncretism with non-adjacent cells. For example, in the last row, the feminine plural form \textit{aAti} maps to four functional feature combinations: \textbf{fp(ad|ac|gd|gc)} -- accusative/genitive and definite/construct.}
\label{suffix-table}
\end{table}

\subsection{Inflection and Cliticization Particularities}
\label{sec:inflec-clitic-particularities}

\paragraph{Default Nominal Suffixes}
The \textbf{default} Arabic nominal suffixes express combinations of four features: gender, number, case, and state. 
As Table~\ref{suffix-table} demonstrates, many of the unique 28 suffixes map to different subsets of the 54 possible feature combinations. Some of the suffixes can be decomposed into smaller compositional units, such as case and state endings with feminine and masculine singular, as well as feminine plural suffixes, but there are some inconsistencies such as the identical accusative and genitive suffixes for feminine plural. While there is a default functional meaning to these morphemes, we find many instances in which there are mismatches between their form and the functional feature values in the word, mostly in number and gender, but also in case and state. We will refer to the morpheme \textbf{forms} using a capitalization of their default \textbf{functional} feature values. For example, \textbf{FP} refers to the suffix set typically associated with the functional features \textbf{fp} without the requirement that the functional features be \textbf{fp}, e.g., \<امتحانات> \textit{AmtHAnAt} `exams' where this is a functionally masculine plural (\textbf{mp}) noun which takes a feminine plural (\textbf{FP}) suffix (see last row in Table \ref{suffix-table}). Taking a \textbf{FP} suffix does not change its functional masculinity. In this case, the function of the \textbf{FP} suffix is not \textbf{fp}, its default, but another value (\textbf{mp}).\footnote{
Some readers may question the logic of the word \<امتحانات> \textit{AmtHAnAt} `exams' being masculine since it requires a feminine number (3-10) quantifier and feminine singular adjective: \<خمسة امتحانات صعبة> \textit{xms{\TAMARBUTA} AmtHAnAt S{\AYN}b{\TAMARBUTA}} `five hard exams'.  However, MSA agreement rules require reverse-gender agreement for number (3-10) quantifiers, and  feminine singular adjective for irrational (non-human) plurals. Furthermore, the singular form \<امتحان> \textit{AmtHAn} `exam' is masculine, and simply pluralizing a noun does not change its gender. For more details, see \newcite{Alkuhlani:2011:corpus}.}

\paragraph{Gender-Number Suffix Mismatch}
\label{par:form-function-mismatch}
Some nominals have suffixes that, by default, express gender and number values that do not match those of the nominals themselves. Examples include \<خليفة> \textit{xaliyfa\TAMARBUTA} `Caliph' (\textbf{ms} noun, \textbf{FS} suffix), 
\<نار> \textit{nAr} `fire' (\textbf{fs} noun, \textbf{MS} suffix),
\<طلبة> \textit{Tlb\TAMARBUTA} `students' (\textbf{mp} noun, \textbf{FS} suffix), and \<نيران> \textit{nyrAn} `fires' (\textbf{fp} noun, \textbf{MS} suffix).

\paragraph{Broken and Other Plurals}
A majority of gender-number suffix mismatches occur with \textbf{broken plurals}, nominals whose number is specified through templatic pattern change. Examples include \<حوامل> \textit{HwAml} `pregnant [p]' (\textbf{fp} noun,  \textbf{MS} suffix),
\<كلاب> \textit{klAb} `dogs' (\textbf{mp} noun, \textbf{MS} suffix), and 
\<طلبة> \textit{Tlb\TAMARBUTA} `students' (\textbf{mp} noun, \textbf{FS} suffix). 
In a minority of cases, there are sound plurals that require slight changes in the stems. An example of such \textbf{semi-sound plurals} is the noun \<حَفَلَات> \textit{HafalaAt} `parties'(\textbf{fp}, \textbf{FP}), whose base stem would suggest the incorrect form \<حَفْلَات>* *\textit{Haf.laAt}. Another case is \textbf{plurals of plurals}, nominals that use broken plural patterns with plural suffixes, e.g.,  \<رِجَالَات> \textit{rijaAlaAt} `leading men'(\textbf{mp} broken plural stem, \textbf{FP} suffix).

\paragraph{Diptotes, Invariables, Indeclinables, and Defectives}
There are many classes of nominals with different variations in terms of how case and state features are realized \cite{Buckley:2004:modern}. 
In contrast to \textbf{triptotes} (the default nominals), \textbf{diptotes} (\<الممنوع من الصرف>), identified typically by pattern or foreign origin, express exceptional syncretism in their case suffixes: \textit{indefinite} diptotes use default definite suffixes, and they also use default accusative suffixes for both accusative and genitive case.  When they are not indefinite, they use default suffixes normally. One example is the noun \<سُفَرَاءَ> \textit{sufaraA'+a} `ambassadors' (\textbf{MSAD} suffix, but ambiguous \textbf{ai}, \textbf{gi}, \textbf{ad}, or \textbf{ac}). 

\textbf{Invariables} use a zero suffix for all case and state features, e.g. \<دُنيا> \textit{dun.yA} `world'.
\textbf{Indeclinables} use the default accusative singular for all cases, e.g., \<فتىً> \textit{fata\AMAQSURA\FATHATAN} `young man'.
And \textbf{Defectives} use the default genitive suffix for nominative in indefinite form, e.g., \<قَاضٍ> \textit{qaAD\KASRATAN} `judge' (\textbf{MSGI} suffix, but ambiguous \textbf{gi}, \textbf{ni}). In addition to the above, there are very special sets of nominals with unique behavior, such as the so-called \textit{five nouns}, which exceptionally represent case in long vowels, e.g., \<أبي> ,\<أبا> ,\<أبو> {\it {\AHAMZAUP}bw}, {\it {\AHAMZAUP}bA}, {\it {\AHAMZAUP}by} `father \textit{of} ...' (nominative, accusative, genitive, respectively). Finally, the  \textbf{MS} suffix (\<اً> \textit{A\FATHATAN}) is written without its \textit{Alif} (long vowel [A]) when the stem ends with a \textit{hamza} (glottal stop), e.g., \<هواءً>  \textit{hwA'\FATHATAN} `air' as opposed to \<هواءاً>*  *\textit{hwA'A\FATHATAN}).

\paragraph{Variable Stem Endings}
\label{par:defective}
There are many nominal classes where the stem ending changes based on the presence of specific suffixes and clitics. 
The following are two of the most common classes. \textbf{Alif-hamza-final} nominal stems vary their \textit{hamza} (glottal stop) form when followed by a clitic. The variation reflects orthographic harmony with the vowels that follow it, e.g., 
\<سُفَرَاءَهُ> \textit{sufaraA\textbf{'a}hu},
 \<سُفَرَاؤُهُ> \textit{sufaraA\textbf{{\WHAMZA}u}hu},
 \<سُفَرَائِهِ> \textit{sufaraA\textbf{{\YHAMZA}i}hi}, `his ambassadors' in accusative, nominative and genitive, respectively.
 \textbf{Defective} nominal stems lose their final letter in some contexts, e.g., \<قَاضٍ> \textit{qaAD\KASRATAN} and \<قَاضِياً> 
 \textit{qaADiyA\FATHATAN}, `a judge' in the nominative/genitive and accusative, respectively. For all such regular cases, we model the varying stem ending as part of the stem buffer.

\paragraph{Proclitics}
Most nominal proclitics do not vary in form when attached to basewords. One common exception is the 
Arabic determiner +\<ال>~\textit{Al+}, whose first letter elides after the prepositional proclitic +\<لِ>~\textit{li+} `for'. The presence of the determiner leads to the addition of a gemination diacritic on the first letter in the baseword  if it is a coronal consonant, aka, \textit{sun letter}, e.g., 
\<شَمْسِ>~+\<ال>~+\<لِ>
\textit{li+Al+{\SHIN}am.si} realizes as \<لِلشَّمْسِ> \textit{lil{\SHIN}{\SHADDA}am.si} `for the sun'.

\paragraph{Enclitics}
\label{par:enclitics}
Pronominal possessive enclitics tend to interact in different ways with stems and suffixes.
Some examples were presented above under \textit{Variable Stem Endings}.
The following are other common cases of such interactions.
The feminine singular suffix \<ة> \textit{\TAMARBUTA} changes to a \<ت> \textit{t} before a clitic, e.g.,
\<نَا>+\<سَفِيرَةُ>
\textit{safiyra{\TAMARBUTA}u+naA} realizes as \<سَفِيرَتُنَا> \textit{safiyratunaA} `our ambassador'.
Similarly, the stem ending \<ى> \textit{ý} turns to \<ا> \textit{A} before a clitic, e.g., \<ي>+\<مَبْنَى> \textit{mab.na{\AMAQSURA}+iy}  \<مَبْنَايَ> \textit{mab.naAya} `my building'.
The 1\textsuperscript{st} person singular pronominal clitic has three allomorphs, and each of the 3\textsuperscript{rd} person pronominal clitics has two.
Table~\ref{tokenization-table}(a) and (b) illustrate one case of the latter (\textit{i+hum}$\rightarrow$\textit{i+him}).

\subsection{Paradigmatic Variation}
\label{sec:paradigm-irregularity}
An important difference between modeling verbal and nominal morphology in Arabic is the  consistent completeness of verbal paradigms (with very few exceptions), and the high degree of variability and incompleteness in nominals. While this issue does not affect the modeling of specific words, it  matters for linking words in the same lexeme and for taming the lexicon. Table~\ref{table:lemma-paradigms} presents examples of different nominal paradigms using a simplified four-slot format covering gender and number (columns) for different lexemes (rows). 
We omit the \textit{dual} value due to its regularity, and  case and state for simplicity. The slots (cells) specify the suffix morphemes using the default values discussed above.

\begin{table}[t!]
\centering
 \includegraphics[width=0.48\textwidth]{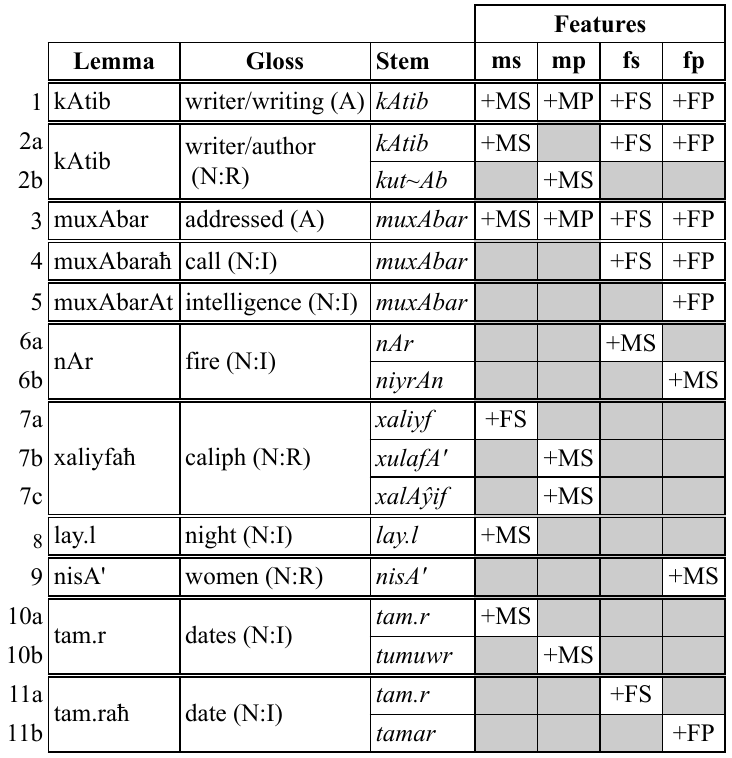}
    \caption{Arabic nominal paradigm examples pairing  \textit{functional} feature values with \textit{form} values. See footnote~\ref{abbreviations} for abbreviations.}
\label{table:lemma-paradigms}
\end{table}

\paragraph{Paradigm Completeness and Stem Count}
A simple standard paradigm uses one stem for all slots and default nominal suffix mapping (perfect match in form and function), e.g., Table~\ref{table:lemma-paradigms}~(1,~3). Some complete paradigms use multiple stems, typically to accommodate one or more broken plurals, e.g., Table~\ref{table:lemma-paradigms}~(2).
%
Incomplete paradigms do not inflect for certain gender and/or number combinations, and some may use one or many stems, e.g., Table~\ref{table:lemma-paradigms}~(all except 1,~2,~3). Of course, some paradigms are complicated by function-form mismatches, e.g., Table~\ref{table:lemma-paradigms}~(6, 7, 9).

\paragraph{Inter-paradigm Ambiguity}
Considering Table~\ref{table:lemma-paradigms}, some paradigm stems seem like they could neatly fit as a subset of a different paradigm, like in the case of Table~\ref{table:lemma-paradigms}~(3, 4, 5), (1 and 2a), and (10 and 11).   However, because they share different meaning spaces and sometimes different POS, they belong to different lexemes.
There is no denying the derivational relationship among these lexemes: they come from the same root and  same initial pattern, but due to derivational specification, the meaning and the paradigm size are affected beyond simple semantic shift. 
For example, lemmas (3, 4, 5) in Table~\ref{table:lemma-paradigms} go from a passive participle adjective (`addressed/called') to a specific common noun (`a call') to a more specific common noun that has no singular (`intelligence services').
The lemma pairs (10 and 11) represent common derivational pairs of mass/collective nouns and instances of them. Given the high degree of variability and inconsistency due to derivational history, this aspect of morphology modeling is complex and demanding.

\section{The {\camelmorph} Approach}
\label{sec:camel-morph-approach}

The {\camelmorph} approach is based on a general framework that could, in principle, be used to build morphological analysis and generation models for any language with concatenative morphology and allomorphic variations \cite{habash-etal-2022-morphotactic}. The {\camelmorph} approach  requires designing \textbf{morphological specifications} describing the language's grammar and lexicon, which are then converted via an offline process powered by its \textbf{DB Maker} algorithm into a \textbf{morphological database} (DB) in the style of BAMA/SAMA DBs \cite{Buckwalter:2004:buckwalter, Graff:2009:standard,Taji:2018:arabic-morphological}. The created DBs can be used by any analysis and generation engine familiar with its format, such as Camel Tools \cite{Obeid:2020:camel}.

The {\camelmorph} morphological specifications can be divided into \textbf{Order} and \textbf{Morpheme} specs. The \textit{order} specifies the positions of all \textit{morpheme classes} in a word. The morpheme class consists of \textit{allomorphs} organized into \textit{morphemes}. These are divided into closed-class (suffixes and clitics), and open-class (stem lexicon) morphemes. 
%
%
Associated with each allomorph is a set of hand-crafted \textbf{conditions}, which control  allomorph selection for a specific morpheme. There are two types of conditions: \textbf{{set} conditions} are activated by the allomorph, and \textbf{{required} conditions} are needed by the allomorph. The lexicon is a large repository that contains the stems and their associated lemmas, and other  features.  Within this framework, the stems also \textit{set} and \textit{require} conditions just like the closed-class morphemes. The offline \textbf{DB Maker} process makes heavy use of these conditions to determine proper combinations and compatibility among the allomorphs in a word. Finally, the framework accommodates the use of ortho-phonological rewrite regex rules (such as sun-letter handling) as part of the analysis/generation engine.

\begin{figure*}[ht!]
\centering
 \includegraphics[width=0.95\textwidth]{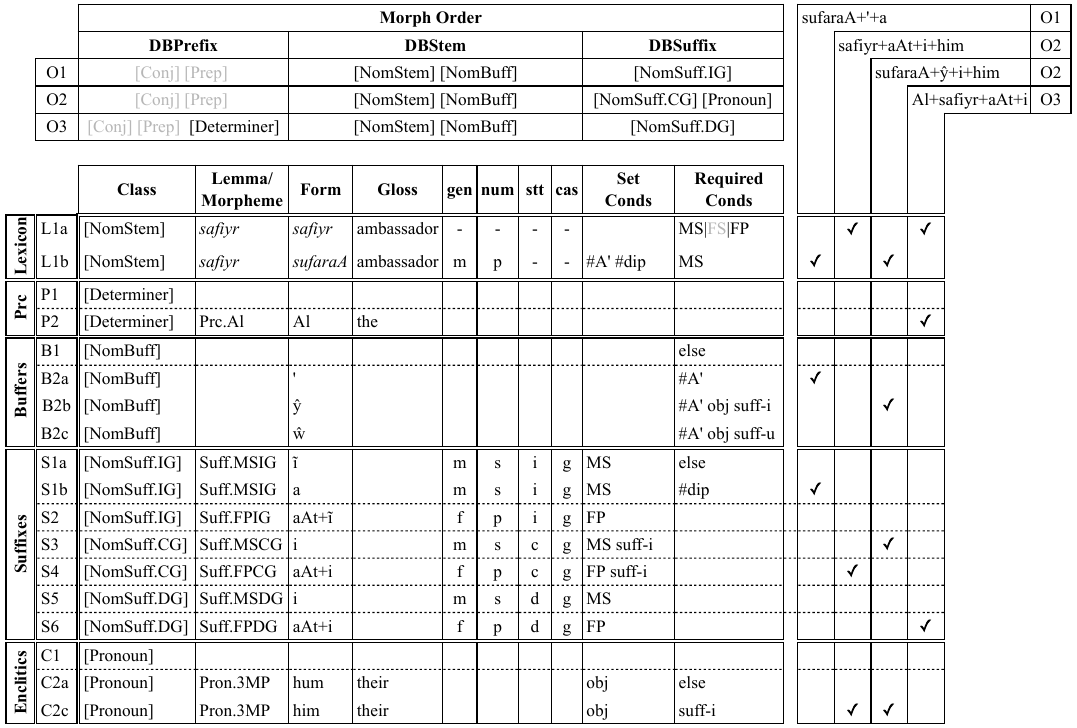}
    \caption{A sample of the {\camelmorph} system implementation for Arabic nominals. The character `|' represents the boolean \textsc{or}, and \texttt{else} represents a negation of the disjunction of conditions below it in the same morpheme. The greyed out elements are not handled in this sample.  See Appendix \ref{sec:conditions-index} for condition meanings.}
\label{lex-order-morph}
\end{figure*}



\section{Modeling Nominals in {\camelmorph}}
\label{sec:nominals-camel-morph}
Next, we discuss the morphological and lexicographic design decisions, which we used to solve all the challenges mentioned in Section \ref{sec:challenges}, and more. The full guidelines will be publicly available (see foonote~\ref{github}).
The last subsection below presents statistics on the resulting database. 

\subsection{Morphotactic Modeling}
Given the complexity of the full system, we employ a highly redacted example 
in Figure~\ref{lex-order-morph} to explain how the system behaves and cover 
the cases in Table~\ref{tokenization-table} and a bit more.

\paragraph{Morph Order}
The top of Figure~\ref{lex-order-morph} shows a segment of the \textbf{Order} part of the \textit{Morphology Specifications} for genitive suffixes.  The order specifies the prepositional clitics that can occur with genitive suffixes, and the relative order of conjunctions, prepositions and determiner clitics (\textbf{DBPrefix}; see also Table~\ref{tokenization-table}).  The stem part consists of a nominal stem and buffer, and the suffix part includes the pronoun enclitic only for the construct suffixes.  The presence of a class in the order sequence does not necessarily mean a morpheme has to be present. Optional classes, such as determiner or pronoun allow a \textit{nothing} option -- see Figure~\ref{lex-order-morph}~(P1,C1).

\paragraph{Lexicon and Buffers}
The \textbf{Lexicon} section shows a lemma with two stems, which together make up a paradigm with a broken plural.
The base stem in Figure~\ref{lex-order-morph}~(L1a) does not specify any feature values as it will acquire them from the suffixes. It lists three required conditions which correspond to the default \textbf{MS}, \textbf{FS} and \textbf{FP} (no \textbf{MP}), as defined in Section \ref{sec:inflec-clitic-particularities}. The broken plural stem (L1b) specifies the gender and number features, which override any features from suffixes. It also indicates being an \textit{Alif-hamza-final} (\texttt{\#A'}) stem and a diptote (\texttt{\#dip}) under \textbf{Set Conditions}, and requires the \textbf{MS} suffix only.
The \textbf{Buffers} section provides the possible segments to complete the \texttt{\#A'} stems under different required conditions.

\paragraph{Suffixes}
The suffixes provided in this redacted example are only for \textbf{MS} and \textbf{FP} (see Section \ref{sec:inflec-clitic-particularities}). Here, we see how a diptote suffix behavior is modeled through the use of the \texttt{\#dip} condition: the morpheme \texttt{Suff.MSIG} has two allomorphs, both of which set the condition \texttt{MS}, but one requires the condition \texttt{\#dip}, and the other requires the negation of \texttt{\#dip} [\texttt{else} of \texttt{\#dip}].
Also, the construct suffixes that interact with pronouns set the condition \texttt{suff-i} indicating the presence of a final /i/. 

\paragraph{Proclitics and Enclitics}
The determiner proclitic in this redacted example has no special constraints. However,
in complete models, the determiner requires that sun letters that follow it take a \textit{shadda} diacritic. Although this requirement is not covered in Figure \ref{lex-order-morph}, it is modeled in our full system with a regex rule in the analysis/generation engine.
The pronoun enclitic, \texttt{Pron.3MP} shows two allomorphs that vary depending on the presence of a suffix /i/, which is set by some of the suffixes.

\paragraph{End-to-End Examples} 
The right-hand side of Figure~\ref{lex-order-morph} demonstrates four cases of morpheme and buffer combinations that this model permits. In essence, the design of the morph class allows all  class members to coexist; but only word forms where all required conditions are actually set are allowed.  For example, the first case of (\<سُفَرَاءَ> \textit{sufaraA+'+a}) uses three elements, which together set the conditions (\texttt{\#A'}, \texttt{\#dip}, \texttt{MS}) and require the same conditions (\texttt{\#A'}, \texttt{\#dip}, \texttt{MS}).
An implausible form such as \<سُفَرَائٍ>* *\textit{sufaraA+{\YHAMZA}+{\KASRATAN}})
would not be allowed as these elements set the conditions (\texttt{\#A'}, \texttt{\#dip}, \texttt{MS}) but require the conditions (\texttt{MS}, \texttt{\#A'}, \texttt{obj}, \texttt{suff-i}, and not \texttt{\#dip}) -- which cannot hold.

Finally, we note that the conditions are agnostic to functional features, and are only concerned with surface form. For example, the lemma \<هَوَاء> \textit{hawaA'} `air' in its functionally masculine singular form would have the stem \<هَوَا> \textit{hawaA}, and set the condition \texttt{\#A'}, the same condition set by the stem \<سُفَرَا> \textit{sufaraA} `ambassadors', which is functionally plural.

\paragraph{Debugging and Quality Check} The space of combinations to validate in the actual system is in the order of billions, of which only a fraction is valid. To debug this system, the generator engine was run on a subset of the nominal paradigm -- chosen along the dimensions which vary the most, using lemmas chosen to represent the continuum of annotated conditions, and the outputs were manually checked by an annotator.

\begin{table}[t!]
\centering
 \includegraphics[width=0.46\textwidth]{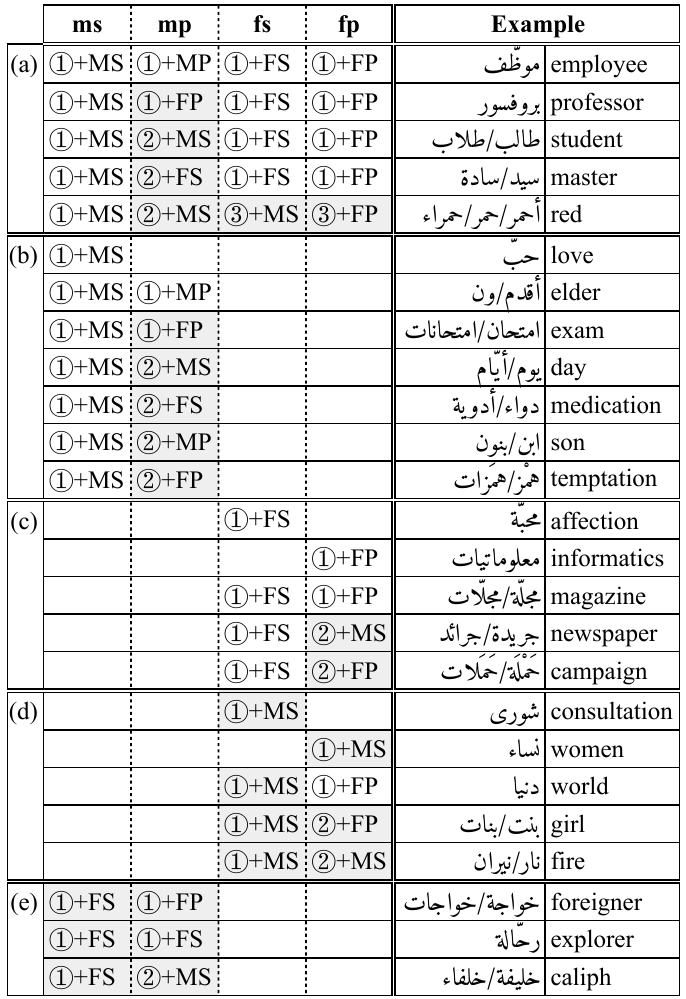}
    \caption{Examples of different lexicographic classes with different degrees of completeness and form-function matching. Greyed out cells mark cases with mismatching form-function in gender or number, or using secondary stems. See Appendix \ref{sec:paradigm-index} for the full table.}
    \vspace{-0.2cm}
\label{table:lexicon}
\end{table}

\subsection{Lexicographic Modeling}
The approach we took to model the morphology of words allows us to clearly disentangle many variables such as case-state, gender-number, and stem class variations. The next step is the lexicographic modeling to group stems belonging to the same lexemes together.
To aid us in modeling the lexicon systematically, we extracted stems and their features from the publicly available CALIMA\textsubscript{\textit{Star}} DB \cite{Taji:2018:arabic-morphological}, and extended its root annotations with patterns, stem paradigms, and lexeme paradigms automatically. With the help of that information, we proceeded to manually annotate (with conditions) and carefully check all the stem clusters (lexemes) for soundness with the help of three annotators. This resulted in all clusters being categorized into one of the lemma paradigms that can be found in Appendix \ref{sec:paradigm-index}. Future lemmas can therefore be added to the lexicon with ease by determining which paradigm they belong to without worrying about conditions. Conditions are only added upon determining the stem paradigm which mainly depends on the surface pattern and form. Were the lexicon conditions not purely concerned with form, it would have not been possible to do that. Therefore the 
{\camelmorph} approach objectively renders the annotators' job simpler as the only layer they are required to interface with is the \textbf{Lexicon}. The annotators should not have to deal with conditions which are internal to the closed-class specifications, i.e., \textbf{Proclitics (Prc)}, \textbf{Buffers}, \textbf{Suffixes}, and \textbf{Enclitics} (see Figure \ref{lex-order-morph}).

As part of this effort, we developed guidelines for making decisions on boundaries between lexemes by 
(a) morpho-syntactic behavior, e.g., agreement patterns and their interaction with rationality \cite{Alkuhlani:2011:corpus}, and (b) semantic change and relationships, e.g., lexical specification turning adjectives into nouns, or systematic derivational relationships between mass/collective nouns and their instance noun forms.   
Given the high degree of variability among nominal lexemes, we developed models for well-formedness checks to identify out-of-norm clusters for quality check.

Table~\ref{table:lexicon} shows 25 lemma paradigms with varying paradigm completeness and gender-number form-function consistency. Circular digits indicate shared stem indices.
\begin{table*}[ht]
\centering
    \includegraphics[width=0.98\textwidth]{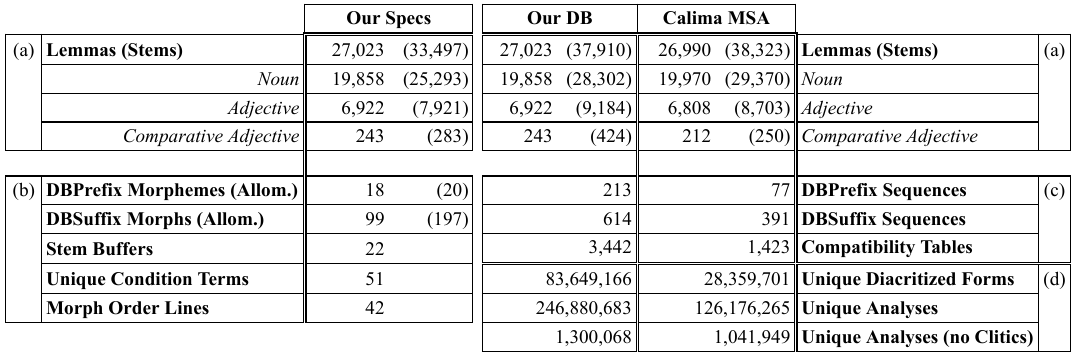}
    \caption{Statistics comparing our morphological specifications and DB with Calima MSA on Arabic nominals.}
\label{table:stats}
\end{table*}

\subsection{Statistics}
\label{sec:statistics}
In this section, we discuss the statistics of our specifications (\textbf{Our Specs}) and their associated resulting DB (\textbf{Our DB}), and we compare \textbf{Our DB} with the \textbf{Calima MSA} DB \cite{Taji:2018:arabic-morphological},\footnote{Version: \texttt{calima-msa-s31\_0.4.2.utf8.db}.} as a baseline, since both have the exact same format. Table \ref{table:stats} contains counts related to the three different entities.

We note that the number of lemmas is the same in \textbf{Our~Specs} and  \textbf{Our~DB}, naturally, and is only slightly larger than \textbf{Calima~MSA}'s. While the number of stems is almost the same in 
\textbf{Our~DB} and \textbf{Calima~MSA}, it is 13\% less in \textbf{Our~Specs} showing that we are able to get comparable results from a more succinct, and hence, more annotator-friendly, way using our morphological modeling. 
Similarly, the small number of morphological modeling elements (Table~\ref{table:stats}.b) and the large number of  complex prefix/suffix sequences they produce (Table~\ref{table:stats}.c) highlight our approach's modeling power.  
The main  reasons for the higher numbers in \textbf{Our~DB} 
in Table \ref{table:stats}.c are
the modeling of the undefined case,\footnote{The \textbf{Calima MSA} model produces a number of analyses with case \textit{undefined} for some suffixes, e.g., \<كِتَابَات> 
 kitaAbaAt `writings' in contrast with defined cases such as  \<كِتَابَاتُ> kitaAbaAtu (see full set in Table~\ref{suffix-table}). However, this treatment is not consistent for all suffixes.  In \textbf{Our~DB}, we extend all suffixes with case \textit{undefined} variants that are in common use.}
%
and the addition of the question proclitic +\<أَ> \textit{{\AHAMZAUP}a}+, which is only present in a few hard-coded cases in  \textbf{Calima~MSA}. These differences translate into \textbf{Our~DB} having roughly two times more analyses than \textbf{Calima~MSA}. The increase is still sensible when clitics are excluded, with an increase of \string~26\% in the analysis count (Table~\ref{table:stats}.d).\footnote{The statistics in Table \ref{table:stats}.d are computed using combinatorics, not generation.}

\section{Evaluation}
\label{sec:evaluation}
We assess the quality of our system by (a) evaluating its coverage of the \textit{training} portion of the Penn Arabic Treebank (PATB; latest versions of parts 1,2,3) \cite{Maamouri:2004:penn} as defined by \citet{Diab:2013:ldc}, and (b) comparing the analyses it generates with those of \textbf{Calima DB} over a list of specific words.

\paragraph{Morphological Coverage} 
For the coverage experiment, we drop all incomplete PATB gold analyses marked with placeholder values ($\sim$1\% of all entries). Of the rest, we are able to recall 95.3\% of gold analyses provided by the PATB (94.5\% in unique type space) based on matching on all of lemma, diacritization, and morphological analysis (BW tag). We performed a human evaluation on a sample of 100 unique words from the mismatching \textit{noun} instances chosen randomly (but weighted by the PATB frequency of the gold analysis). We found that 86\% of mismatches are due to gold inconsistencies or errors. These include 
-- among other issues listed in Section \ref{sec:paradigm-irregularity} -- 
spelling inconsistencies between lemma and stem, or attributing a stem to a wrong lemma because of paradigm ambiguity. Our system produces valid analyses for these cases, but it fails for the remaining 14\%. 
A similar 100 adjective sample reveals that 95\% of mismatches are due to  inconsistent gold tags, and are mainly due to a wrong POS attribution and lemma-stem spelling mismatch. Our system handles these cases correctly. In the released version, we made sure to include all missing analyses.

\paragraph{Analysis Evaluation} 
Finally, we choose 50 random words from the 100-sample taken for the nouns in the previous paragraph for closer inspection, and we manually compared all analyses generated by both \textbf{Our~DB} and \textbf{Calima DB} for these words.
Of the union of all manually inspected analyses generated by the two systems (1,406 analyses for the 50 words), 21\% are generated by both, 44\% are generated only by \textbf{Our~DB}, and 35\% are generated only by \textbf{Calima DB}. 
We find that about 60\% of the analyses generated only by \textbf{Our~DB} are due to unmodeled or incompletely modeled phenomena in \textbf{Calima DB}, e.g., the question proclitic morpheme
or some instances of the \textit{undefined} 
case. The remaining 40\% are due to inaccurate modeling on the \textbf{Calima DB} side. 
For example, 
\textbf{Calima DB} only provides one lemma for \<معلومات> \textit{ma{\AYN}.luwmaAt}, \<معلوم> \textit{ma{\AYN}.luwm} `known',  and misses the lemma \<معلومة> \textit{ma{\AYN}.luwma\TAMARBUTA} `a piece of information', while \textbf{Our~DB} provides both.

One systematic mistake is
 allowing the +\<ال>~\textit{Al+} determiner to attach to construct noun stems, whereas this behavior should only be restricted to adjectives participating in a 
 \textit{False Idafa} construction
 (\<إضافة لفظية>), e.g., \<الأبيض اللّون> `the-white-colored'  
 \cite{hawwari-etal:2016:explicit}.
 Other mistakes include wrong lemma gender, and spelling inconsistencies between lemma and stem.  Finally, about 6\% of the \textbf{Our~DB} analyses in  this sample are admittedly wrong, but can easily be fixed in our specifications.

\section{Conclusion and Future Work}

We presented a detailed review of the challenges of modeling Arabic nominals morphologically and lexically. We developed an annotator-friendly and easily extendable system 
for modeling nouns, adjectives and comparative adjectives building on an existing open-source framework for Arabic morphology. We evaluated our system 
against a popular analyzer for Arabic, showing that our resulting database is more consistent and provides a more accurate linguistic representation. We make our models, system details, and guidelines publicly available (see footnote~\ref{github}).

In the future, we plan to extend our work to other MSA POS tags and to Arabic dialects.
We also plan to make our model more robust to spelling variations and integrate it in downstream applications, e.g., morphological disambiguation, tokenization and diacritization \cite{obeid-etal-2022-camelira}, readability visualization \cite{hazim-etal-2022-arabic}, gender rewriting \cite{alhafni-etal-2022-user}, error typing \cite{belkebir-habash-2021-automatic}, and grammatical error correction \cite{alhafni-etal-2023-advancements}.
 

\section{Limitations}

The current system faces several limitations: it lacks robustness in handling input orthographic errors, restricting its usability in spontaneous orthography contexts. Additionally, it does not comprehensively model valid spelling variants commonly used. The high coverage generates numerous options, including some less likely but theoretically correct ones, potentially overwhelming downstream processes without optimized filtering and ranking models. There is also a lack of explicit linking across lemmas sharing derivational history. Furthermore, the model is currently limited to nouns, adjectives, and comparative adjectives, representing the open-class nominals at this stage.

\section{Ethics Statement}
All annotators received fair wages for their contributions to the development, quality checking of lexical resources, and debugging the overall system. While we recognize the possibility of unforeseen errors in our lexical resources, we anticipate that the associated risks to downstream applications are minimal. Additionally, we acknowledge that, like many other tools in natural language processing, our tool could be misused in the wrong hands for manipulating texts for harmful purposes.



\bibliography{anthology,camel-bib-v3,custom}

\appendix
\onecolumn
\clearpage

\section{Glossary of Terms and Abbreviations}
\label{glossary}

\begin{table*}[h!]
\centering
 \includegraphics[width=0.85\textwidth]{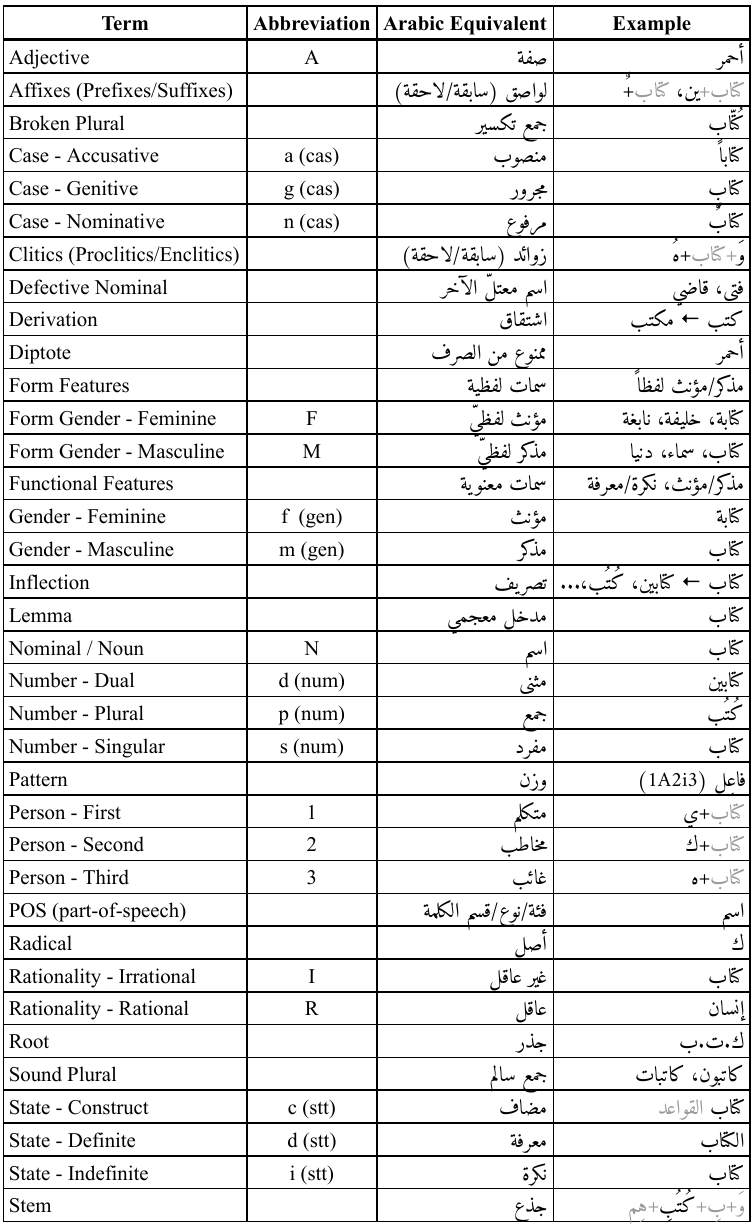}
    \caption{Table featuring the Arabic equivalents of the terms used in this paper, including their abbreviations.}
\label{table:glossary}
\end{table*}

\clearpage

\section{Conditions Index}
\label{sec:conditions-index}

\begin{table*}[h!]
\centering
 \includegraphics[width=\textwidth]{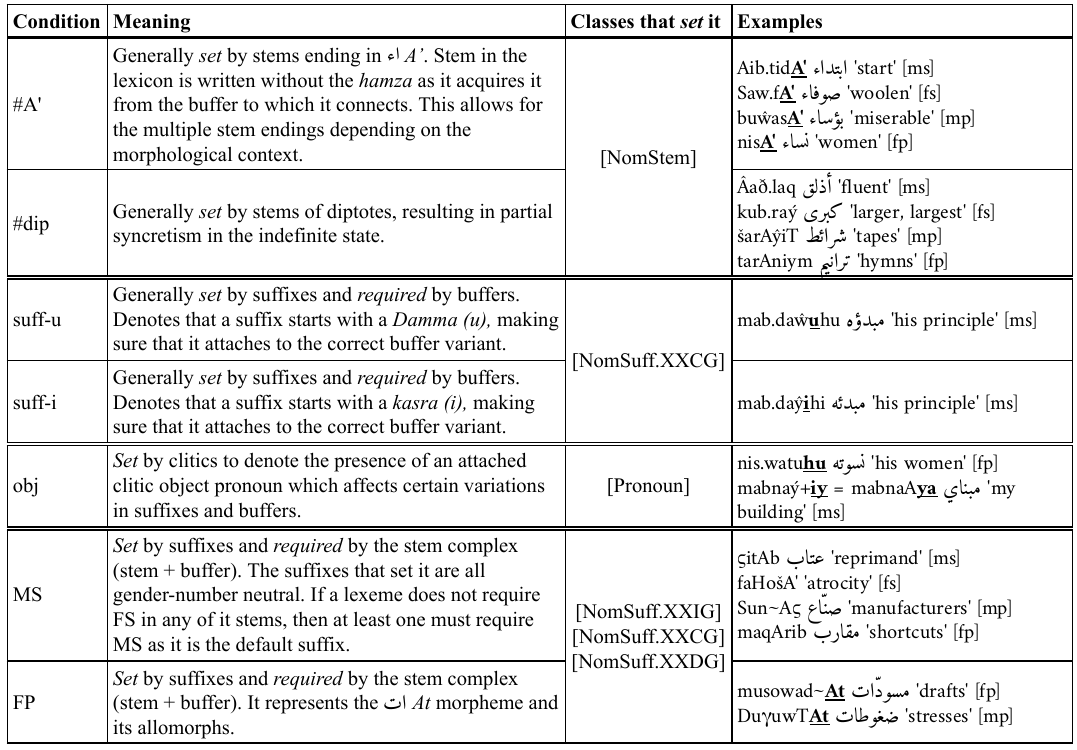}
    \caption{Index of pre-defined conditions used in Figure \ref{lex-order-morph} and their meanings, with examples.}
\end{table*}

\clearpage

\section{Nominal Lemmas Paradigm Index}
\label{sec:paradigm-index}

\begin{table*}[h!]
\centering
 \includegraphics[width=\textwidth]{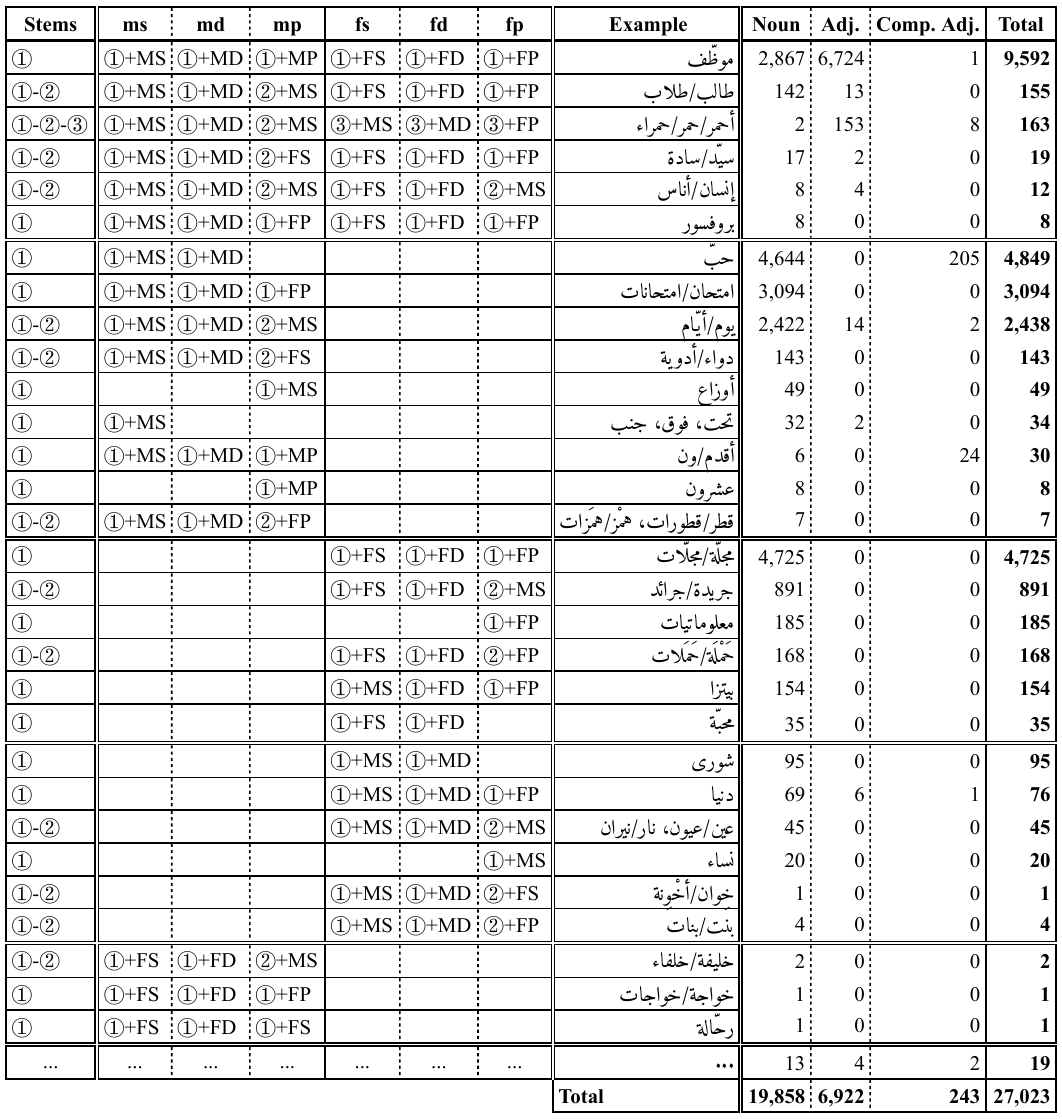}
    \caption{Index of basic lemma paradigms identified. See Appendix \ref{glossary} for abbreviations and Section \ref{sec:paradigm-irregularity} for an explanation of the form feature suffix sets. Statistics included pertain to the number of lemmas per paradigm for each POS.}
\label{table:paradigm-index}
\end{table*}

\end{document}

%% file: macros.tex
\newcommand{\hide}[1]{}

\newcommand{\AHAMZAUP}{{\^{A}}}
\newcommand{\WHAMZA}{{\^{w}}}

\newcommand{\YHAMZA}{{\^{y}}}
\newcommand{\TAMARBUTA}{{$\hbar$}}

\newcommand{\SHIN}{{\v{s}}}
\newcommand{\AYN}{{$\varsigma$}}

\newcommand{\AMAQSURA}{{\'{y}}}

\newcommand{\FATHATAN}{{\~{a}}}
\newcommand{\KASRATAN}{{\~{\i}}}

\newcommand{\SHADDA}{{$\sim$}}